\documentclass[letterpaper, 10 pt, conference]{ieeeconf}  

\IEEEoverridecommandlockouts                              

\usepackage{graphics} 
\usepackage{epsfig} 
\usepackage{mathptmx} 
\usepackage{times} 
\usepackage{amsmath} 
\usepackage{amssymb}  
\usepackage{xcolor}
\usepackage{booktabs}
\usepackage{multirow}

\def\etc{\emph{etc. }}

\title{\LARGE \bf
Open-Nav: Exploring Zero-Shot Vision-and-Language Navigation in Continuous Environment with Open-Source LLMs
}

\author{Yanyuan Qiao$^{1}$, Wenqi Lyu$^{1}$, Hui Wang$^{1}$, Zixu Wang$^{2}$, Zerui Li$^{1}$, Yuan Zhang$^{1}$, Mingkui Tan$^{2}$, Qi Wu$^{1*}$
\thanks{1 The authors are with the Australian Institute for Machine Learning at the University of Adelaide, Adelaide, Australia .}%
\thanks{2 Zixu Wang and Mingkui Tan are with the School of Software Engineering, South China University of Technology, Guangzhou, China.}%
\thanks{*Corresponding author: Qi Wu ({\tt\footnotesize qi.wu01@adelaide.edu.au})}
}

\begin{document}

\maketitle
\thispagestyle{empty}
\pagestyle{empty}

\begin{abstract}
Vision-and-Language Navigation (VLN) tasks require an agent to follow textual instructions to navigate through 3D environments. Traditional approaches use supervised learning methods, relying heavily on domain-specific datasets to train VLN models. Recent methods try to utilize closed-source large language models (LLMs) like GPT-4 to solve VLN tasks in zero-shot manners, but face challenges related to expensive token costs and potential data breaches in real-world applications. In this work, we introduce Open-Nav, a novel study that explores \textit{open-source} LLMs for zero-shot VLN in the \textit{continuous} environment. Open-Nav employs a spatial-temporal chain-of-thought (CoT) reasoning approach to break down tasks into instruction comprehension, progress estimation, and decision-making. It enhances scene perceptions with fine-grained object and spatial knowledge to improve LLM's reasoning in navigation. Our extensive experiments in both simulated and real-world environments demonstrate that Open-Nav achieves competitive performance compared to using closed-source LLMs. The project webpage is https://sites.google.com/view/opennav
\end{abstract}


\section{INTRODUCTION}
The Vision-and-Language Navigation (VLN) task~\cite{r2r} has garnered significant attention in recent years. In this task, an agent follows textual instructions to navigate within a 3D indoor environment, particularly in unseen rooms.
Early research idealized the task setting by focusing on discrete configurations, where navigation is simplified to traversing a predefined graph of the environment. These predefined graphs introduced inherent assumptions about navigation positions and perfect movement between nodes, limiting the task's applicability to real-world scenarios. 
To address these limitations and more closely mimic real-world navigation conditions, the VLN in continuous environments (VLN-CE)~\cite{Krantz2021WaypointMF, vlnce} setting was introduced. Unlike its predecessor, VLN-CE does not rely on predefined graphs, allowing more flexibility and realism in navigating dynamic environments.

Initial VLN methods predominantly employed LSTM-based architectures, exploring various research directions such as auxiliary tasks~\cite{regretful,Zhu0CL20}, reinforcement learning strategies~\cite{rcm}, and data augmentation~\cite{Li2022EnveditEE,wang2023scaling}, \etc. However, with the advent of Transformer-based models, the focus shifted towards leveraging transformers for enhanced representation learning~\cite{airbert} and the integration of historical information~\cite{hamt}. More recent efforts have involved fine-tuning large models like NaviLLM~\cite{navillm} and Navid~\cite{Zhang2024NaVidVV} for VLN tasks.

These learning-based VLN methods rely heavily on large-scale, domain-specific datasets annotated with expert instructions. Consequently, they are limited by their dependence on extensive training data and face challenges such as the simulation-to-reality gap, where agents trained in simulated environments struggle to generalize to real-world settings.

Inspired by the development of large language models (LLMs), recent works have explored zero-shot agents powered by large language models (LLMs), which serve as the “brains” of embodied agents. Notably, SayCan\cite{saycan} demonstrated how LLMs could extract and apply knowledge to complete physics-based tasks, including simple navigation and object manipulation. For VLN tasks, several approaches have been explored using LLMs as navigators in a zero-shot, training-free manner. NavGPT\cite{Zhou2023NavGPTER} was one of the first to use GPT-4 as a zero-shot navigator, converting visual observations of candidate viewpoints into textual descriptions, which were then processed by the LLM to decide the next action. Similarly, DiscussNav \cite{Long2023DiscussBM} introduced specialized roles where ChatGPT handled instruction analysis, InstructBLIP \cite{Dai2023InstructBLIPTG} provided visual perception, and GPT-4 performed completion estimation and decision testing. This collaborative framework allowed for a modular and explainable decision-making process. An additional advantage of LLM-based approaches lies in their transparency. Traditional VLN models often function as black boxes, making it difficult to understand the agent’s reasoning behind specific action predictions. In contrast, LLMs can articulate their decision-making process, providing greater insight into their thought process while making navigation decisions.

\begin{figure}[t]
   \centering
   \includegraphics[width=1\linewidth]{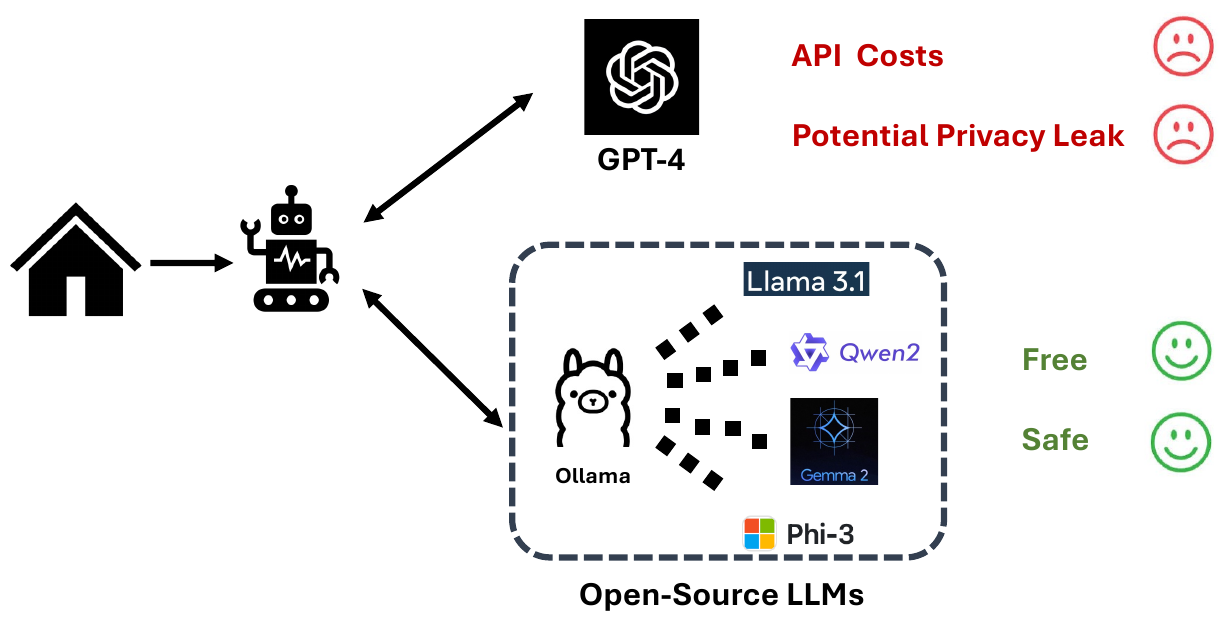}
   \vspace{-15pt}
      \caption{Comparison between GPT-based Navigator and open-source LLM-based Navigator. The GPT-based Navigator requires continuous queries to the GPT model via API for navigation, incurring high costs and necessitating the transmission of environmental data to servers, which raises privacy concerns. In contrast, the open-source LLM-based Navigator utilizes locally deployed LLMs, which are not only free but also safeguard user privacy by eliminating the need to transmit sensitive data.}
    \label{fig:teaser}
    \vspace{-15pt}
\end{figure}

While these approaches have shown the feasibility of using LLMs as navigators, they also present significant challenges, especially in real-world scenarios. 
First, methods like NavGPT and DiscussNav rely heavily on API calls to advanced models such as GPT-4, which results in substantial operational costs. 
Additionally, in real-world scenarios like home robotics or industry robotics, users may be sensitive to transmitting data about their indoor environments to external servers due to privacy concerns, as this involves sharing private information with remote models.

Furthermore, these above LLM-based VLN methods are primarily implemented in discrete environments, where fewer factors must be considered. For example, visual perception is often simplified by converting RGB images into textual descriptions. However, in continuous environments, the navigator must consider additional factors, such as the distance to a target, potential obstacles, and collision risks. These factors are crucial for navigation in complex, real-world scenarios.

To address these challenges, we introduce Open-Nav, an empirical study that explores open-source LLMs for zero-shot Vision-and-Language Navigation in Continuous Environments.
\textbf{\textit{First}}, rather than relying on costly API-based models like GPT-4, we deploy open-source LLMs locally. Using the Ollama framework, we evaluated four recent state-of-the-art open-source LLMs: Llama3.1-70B, Qwen2-72B, Gemma2-27B and Phi3-14B as navigators. Figure~\ref{fig:teaser} illustrates the comparison of closed-source GPT-based navigators and locally-deployed open-source LLM-based navigators. To enable more effective reasoning, we adopted a spatial-temporal chain-of-thought (CoT)~\cite{cot} approach, which allows the LLMs to break down complex navigation tasks into multiple reasoning stages: instruction comprehension, progress estimation, and decision-making. This multi-stage framework ensures that the LLM makes more accurate action predictions. 
\textbf{\textit{Second}}, to facilitate the transition of existing LLM-based agent methods to continuous environments, we retained the traditional waypoint predictor used in VLN-CE tasks to identify navigable points in space, which is more efficient than directly predicting low-level actions~\cite{Krantz2021WaypointMF,dcvln}. 
\textbf{\textit{Third}}, to narrow the gap between closed-source LLMs and open-source models, we enhanced the agent’s visual perception by improving the quality of environment descriptions with spatial and fine-grained object knowledge. We incorporated both RGB and depth data from the current visual observations, allowing the LLM to gain a better sense of spatial relationships. For spatial perception, we employ a spatial understanding Vision-Language-Model (VLM) SpatialBot~\cite{Cai2024SpatialBotPS}, which describes objects within the scene and provides spatial information, such as distances between objects and the agent from both depth and RGB observations. For detailed object recognition, we utilized the fine-grained object recognition model RAM~\cite{Zhang2023RecognizeAA} to detect and identify specific objects.
We perform comprehensive experiments to assess the effectiveness of our Open-Nav method in both simulated and real-world environments, validating its performance across diverse settings.

In this work, our main contributions are:
\begin{itemize}
    \item We propose Open-Nav, a novel study that explores open-source LLMs for zero-shot VLN-CE, reducing costs and addressing privacy concerns by eliminating the need for closed-source LLMs like GPT-4.
    \item We adopt a spatial-temporal chain-of-thought approach for effective reasoning, which incorporates both RGB and depth data to enhance visual perception and spatial reasoning for LLM navigator.
    \item We perform extensive experiments in simulated and real-world environments, demonstrating that Open-Nav achieves competitive performance compared to using closed-source LLMs.
\end{itemize}

\vspace{-5pt}
\section{{Related Work}}
\subsection{Vision-and-Language Navigation}
The Vision-and-Language Navigation (VLN) task requires an agent to follow verbal instructions and visual cues to reach a target location~\cite{Dorbala2022CLIPNavUC}. Early approaches predominantly used a discrete VLN setup with the MP3D simulator~\cite{mattport}, where agents navigated through predefined graphs, focusing on high-level decisions while ignoring fine-grained movements. More recently, the focus has shifted to continuous environments~\cite{vlnce} using the Habitat simulator, where agents perform basic actions like moving forward or rotating, providing a more realistic navigation experience.

Early approaches were based on an encoder-decoder framework~\cite{r2r}, which employed sequence-to-sequence architectures built on LSTM and attention mechanisms. Subsequent research aimed to improve VLN performance from various directions, including enhanced representation learning~\cite{oaam, nvem, graph}, reinforcement learning~\cite{rcm}, and data augmentation~\cite{Li2022EnveditEE,wang2023scaling,
speakerfollower,envdrop}. There was also a notable focus on integrating external knowledge sources to improve generalization~\cite{li2023kerm}.
With the rise of Transformer architectures~\cite{dcvln}, many works introduced pre-training methods specifically tailored for VLN tasks~\cite{airbert,prevalent,altr,bertvln, qiao2023hop+}. 
In addition, DUET~\cite{Chen_2022_DUET} and ETPNav~\cite{An2023ETPNavET} constructed topological maps to capture global navigation information, while BEVBert~\cite{an2023bevbert} generated top-down semantic maps, and GridMM~\cite{wang2023gridmm} introduced a dynamically growing top-down egocentric grid memory map.
In addition to architectural advancements, some recent works have fine-tuned large models for VLN tasks. For example, NaviLLM~\cite{navillm} modified the Vicuna model to enable diverse embodied tasks, while Navid~\cite{Zhang2024NaVidVV} collected 550k navigation samples and fine-tuned Vicuna to create a vision-language model (VLM) specifically for VLN tasks.
Different from these works, we try to explore training-free VLN in continuous environments in a zero-shot manner with open-source LLMs as navigators.

\subsection{Large Language Models for Navigation}
Benefiting from the rise of large language models, recent studies have explored the application of LLM in VLN tasks.
Some studies use LLM as a planner, such as LLM-Planner~\cite{song2022llm} generates detailed plans consisting of sub-goals that are dynamically updated by incorporating detected objects and following predefined programming rules. Similarly, methods such as MiC~\cite{qiao2023march} and $A^2$Nav~\cite{a2vnav} focus on decomposing navigation tasks into detailed text instructions. MiCs provides step-by-step plans from static and dynamic perspectives, while $A^2$Nav uses GPT-3 to parse instructions into actionable sub-tasks. These methods mainly use LLM as a knowledge generator, while action decisions are still within the purview of navigators trained through supervised learning.
There are also some works that directly use LLM as a navigator, usually visual observations are converted into text descriptions and input into LLM together with instructions, and then LLM performs action prediction.
NavGPT was the first to demonstrate the feasibility of zero-shot navigation, and NavGPT~\cite{Zhou2023NavGPTER} uses GPT-4 to autonomously generate actions. DiscussNav~\cite{Long2023DiscussBM} extends this approach by deploying multiple domain-specific LLM experts to automate and reduce human involvement in navigation tasks. 
MC-GPT \cite{zhan2024mc} employs memory topology graphs and human navigation examples to diversify strategies, while InstructNav~\cite{Long2024InstructNavZS} decomposes navigation into subtasks with multi-source value graphs for efficient execution.
Most of the previous work is based on GPT4, but calling GPT4 requires repeated API access and is expensive, and sending indoor scene information to the cloud will involve user privacy issues. 
In this paper, we built a VLN pipeline, the main purpose of which is to use the open-source LLM as a navigator, we also made some designs to enhance spatial perception. Specifically, for the scene understanding, we integrated the spatial position relationship and the distance between the object and the agent.

\begin{figure*}[t]
      \centering
      \includegraphics[width=1\textwidth]{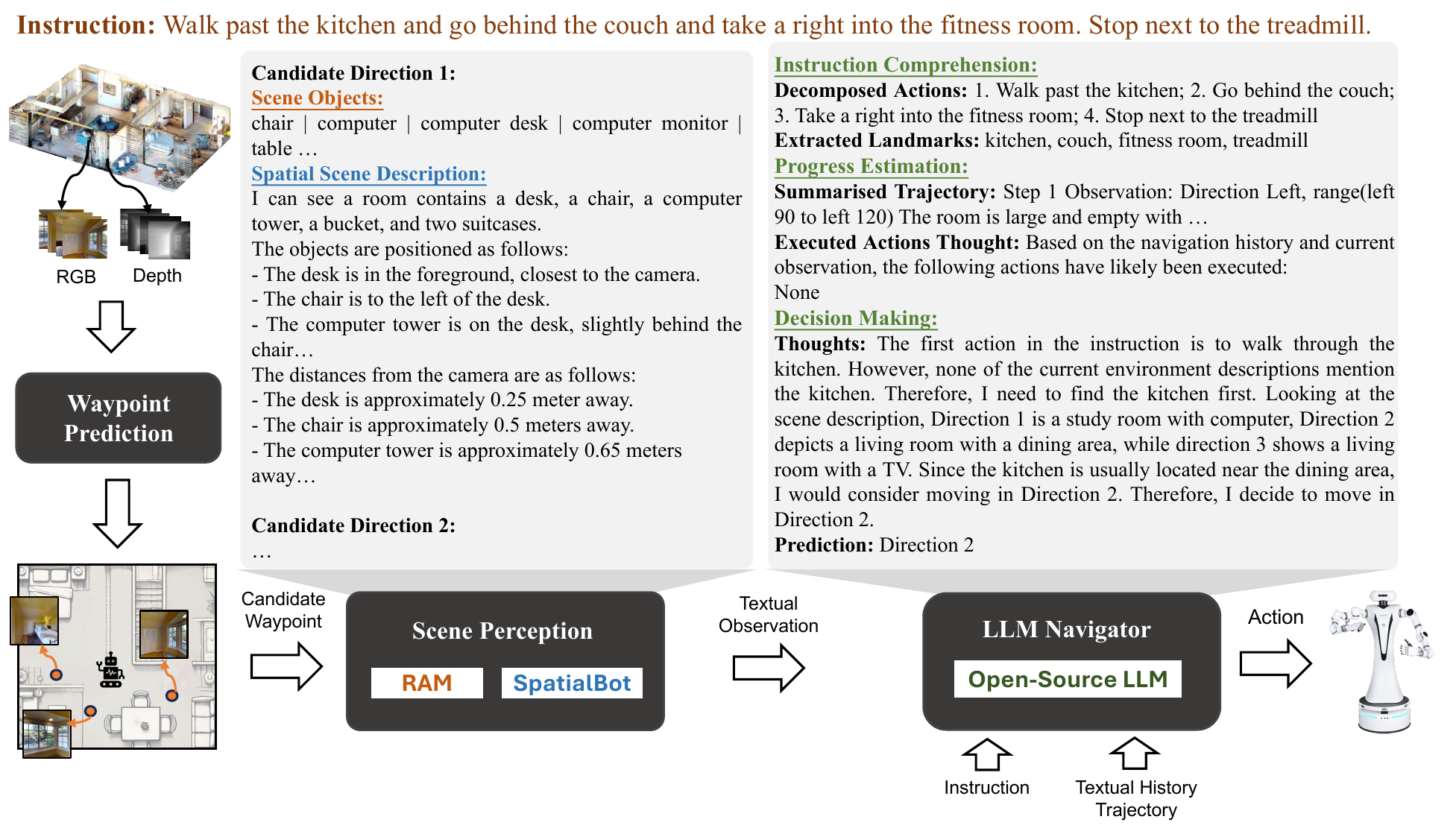}
      \vspace{-20pt}
       \caption{Overview of Open-Nav. The Waypoint Prediction module uses panoramic RGB and depth images to pinpoint potential navigation waypoints, which are then analyzed by the Scene Perception module. This module processes these images to determine object locations and spatial relationships and recognize scene elements. This data is then converted into a text format. The LLM Navigator performs tasks in three stages: understanding instructions, estimating progress, and making decisions. Finally, the navigator determines the precise location to navigate to and performs the corresponding actions.}
      \label{fig:overview}
      \vspace{-10pt}
\end{figure*}

\section{{PROBLEM FORMULATION}}
Vision-and-Language Navigation in Continuous Environments (VLN-CE) is a task that requires an autonomous agent to navigate a 3D mesh using verbal instructions to follow a specified path and reach a designated target location. The agent performs low-level actions, such as moving direction and distance.
At each position,  the agent obtains panoramic RGB visuals \(O = \{r_{gb}, r_d\}\), comprised of 12 RGB and 12 depth snapshots taken at regular intervals across a 360° arc (0°, 30°, ..., 330°). Additionally, the agent is provided with verbal directives for each trial, with the linguistic embeddings of these instructions represented as \(W = \{w_i\}_{i=1}^L\), where \(L\) denotes the number of words in each instruction.

\section{Framework Design}
The proposed framework is shown in the Figure~\ref{fig:overview}. The Waypoint Prediction module identifies potential navigation points from panoramic RGB and depth images. Then, the Scene Perception module processes these potential navigation points to identify objects and extract object positions and spatial relationships in the scene. This spatial information is integrated into textual format observations. At last, for the given instruction and textualized observations, LLM Navigator will complete the task in three stages: instruction comprehension, progress estimation, and decision-making. Finally, the navigator selects the specific location to go to and then performs the action.

\subsection{Waypoint Prediction}
We employ a transformer-based model for predicting waypoints by integrating RGB and depth image features using two specialized ResNet50~\cite{resnet}networks: one for RGB and another for depth data.
At each evaluated location, feature vectors for RGB and depth images are extracted and fused:

\begin{equation}
v_{i}^{\text{rgbd}} = W_m (f_{\text{ResNet-RGB}}(I_{i}^{\text{rgb}}) \parallel f_{\text{ResNet-Depth}}(I_{i}^{d}))
\end{equation}

These fused vectors $v_{i}^{\text{rgbd}}$ are processed by a two-layer transformer network to model spatial relationships between views. The network's self-attention is restricted to adjacent views to enhance spatial reasoning:

\begin{equation}
\tilde{v}_{i}^{\text{rgbd}} = \text{Transformer}(\{v_{i-1}^{\text{rgbd}}, v_{i}^{\text{rgbd}}, v_{i+1}^{\text{rgbd}}\})
\end{equation}

The output from the transformer is used to generate a heatmap of potential waypoints, which is then refined through non-maximum suppression (NMS) to focus on the most probable locations:

\begin{equation}
H_{\text{refined}} = \text{NMS}(\text{MLP}(\tilde{v}_{i}^{\text{rgbd}}))
\end{equation}

From this refined heatmap, $K$ nearby waypoints $\Delta W = \{\Delta w_i\}_{i=1}^{K}$ are selected, where each waypoint $\Delta w_i$ is defined by a pair of values: the angle in radians and the distance in meters from the agent, providing precise directional and spatial information necessary for navigation.

\subsection{Scene Perception}
Traditional approaches in scene understanding for VLN tasks often rely on captioning models such as BLIP~\cite{Li2022BLIPBL} and InstructBLIP~\cite{Dai2023InstructBLIPTG} to generate descriptions from RGB images. These descriptions generally include room types, furniture, and other common objects but lack detailed spatial relationships and distances between objects.

To address these limitations, we employ the spatial understanding VLM SpatialBot~\cite{Cai2024SpatialBotPS} and fine-grained object recognition model RAM~\cite{Zhang2023RecognizeAA} in a zero-shot manner to provide detailed spatial information about candidate waypoints. SpatialBot processes RGB and depth images to create enriched scene descriptions that include spatial information. The process involves combining RGB and depth images into a unified tensor after adjusting their aspect ratios if necessary. The processed images are then fed into SpatialBot along with a textual prompt that guides the generation of descriptions emphasizing spatial relationships and distances:
\begin{equation}
D_\text{spatial} = \text{SpatialBot}([I_{rgb}, I_{d}])
\end{equation}

Simultaneously, we utilize the RAM model for fine-grained object detection and recognition using RGB images. RAM identifies and classifies objects within the scene, providing their spatial coordinates and labels. 
Let \( \mathcal{O} = \{o_1, o_2, \dots, o_n\} \) denote the set of objects detected in the scene.
The detection process involves analyzing the RGB image to locate objects and determine their positions in a three-dimensional space:
 \begin{equation}
o_i = \texttt{RAM}(I_{rgb}) \quad \text{for} \quad i = 1, 2, \dots, n
 \end{equation}

\subsection{Spatial-Temporal Chain-of-Thought for LLM Navigator}
Given the instruction $I$, textualized spatial observations $O_{text}=(D_\text{spatial},\{o_i\})$ and textualized history trajectory $T_{text}$ summarized by the LLM navigator, we elaborately design a comprehensive spatial-temporal chain-of-thought from three viewpoints to enhance the LLM's ability in navigation:
instruction comprehension, navigation progress estimation, and action decision-making.

\subsubsection{Instruction Comprehension}
To enable effective navigation, the LLM is tasked with breaking down the given navigation instructions into two key components: decomposed actions and extracted landmarks. Decomposed actions refer to the steps that the agent must execute during navigation (e.g., ``turn left'', ``move forward''), while extracted landmarks refer to objects or locations in the environment that guide the agent (e.g., ``the door'', ``stairs''). We designed specific prompts for each instruction:
\begin{quote}
``\textit{You are an action decomposition expert. Your task is to detect all actions/landmarks in the given navigation instruction. Ensure the integrity of each action/landmark. Your answer must consist ONLY of a series of labeled action phrases without beginning sentences. Can you decompose actions/landmarks in the instruction? Actions:}''
\end{quote}

\subsubsection{Progress Estimation}
To effectively monitor and evaluate the agent's navigation progress, we utilize a comprehensive review of the navigation history, which incorporates summaries of past visual observations alongside the agent's reasoning processes. The progress estimation is conducted through a meticulously structured analysis comprising the following steps:

\begin{itemize}
    \item {Landmark and Action Verification:} Identify which decomposed landmarks or actions have been encountered or executed based on navigation history.
    \item {Directional Analysis:} Assess changes in directions at each navigational step to ensure that the agent adheres to the prescribed route.
    \item {Action Completion Estimation:} Compare each action stipulated in the instructions against the agent’s movement and decision-making history to determine their completion status.
    \item {Sequential Evaluation:} Enforce a sequential checking mechanism where the completion of subsequent actions is contingent on the completion of predecessors.
\end{itemize}

This analytical framework enables the LLM to effectively track the agent’s progress by determining what portions of the navigation task have been completed and what remains. Such an approach not only facilitates the evaluation of navigation success but also provides crucial feedback on whether the agent is accurately following the given instructions.

\subsubsection{Decision Making}
Before the LLM makes an action decision, we integrate the key information as the spatial-temporal chain-of-thought to enhance the LLM's decision-making process. This information includes the available candidate viewpoints that the agent can choose from at the current step, the specific instruction being followed and its decomposed actions and landmarks, a summary of the agent’s past actions and decisions to provide historical context, an estimate of which actions have been completed so far, and observations from the agent’s current position, including a visual description of each candidate waypoint including spatial knowledge. At last, the LLM evaluates the spatial and navigation environment, generates the decision thought, and selects the most appropriate candidate viewpoint.

\section{Experiments}

\subsection{Experiment Setup}
\noindent\textbf{Simulated Environments.}
We evaluate our approach using the R2R-CE dataset within simulated environments. R2R-CE is based on the Matterport3D~\cite{mattport} scenes, and it converts the discrete paths of the original R2R dataset into continuous environments using the Habitat Simulator. 
Following the setting in previous LLM-related works~\cite{Zhang2024NaVidVV}, we randomly sample a subset of about 100 episodes.
In the simulated evaluation, the model was run on a single RTX 3090 GPU.

\noindent\textbf{Real-world Environments.}
To further evaluate our approach in real-world settings, we designed comprehensive experiments based on the following methods~\cite{Zhang2024NaVidVV}. These experiments cover a range of indoor environments with different spatial complexity. We selected three different indoor scenes: {office, laboratory, and game room. }
For each layout, we annotated 20 instructions, including both simple and complex instructions. This combination allows us to evaluate the agent's ability to handle basic and advanced navigation scenarios in real-world environments.
In the real-world evaluation, the VLN model was run on a laptop equipped with a GeForce RTX 3080 GPU.

\begin{figure}[h]
   \centering
   \vspace{-10pt}
     \includegraphics[width=1\linewidth]{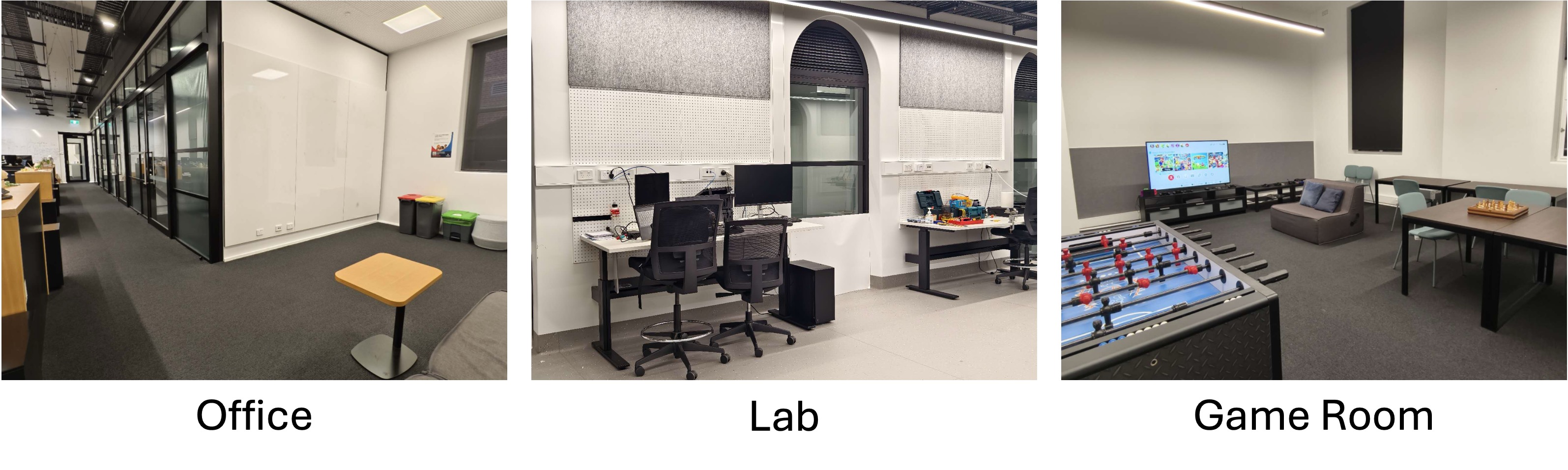}
     \vspace{-22pt}
    \caption{The arrangement of the real-world environment.}
    \vspace{-10pt}
    \label{fig:vis}
\end{figure}

\noindent\textbf{Implementation Details}
For simulated experiments, we utilize the waypoint predictor pretrained in~\cite{dcvln}. %
For real-world experiments, we used a Dual-Arm Composite Robot equipped with a Realsense D435 camera. 
The RGB image dimensions are 244×244×3 and the depth dimensions are 256×256.
For LLM navigator, we use Ollama to deploy 4 open-source LLMs: Llama3.1-70B-instruct, Qwen2-72B-instruct, Gemma2-27B-instruct and Phi3-14B-instruct. To run these LLMs on a single RTX A6000 GPU with 48GB memory, we use the 4bit-quantized models of Llama3.1-70B-instruct and Qwen2-72B-instruct, 8bit-quantized model of Phi3-14B-instruct, and FP16 model of Phi3-14B-instruct.

\noindent\textbf{Evaluation Metrics}
We evaluate the agent’s navigation performance using standard VLN metrics, including success rate (SR), oracle success rate (OSR), normalized Dynamic Time Warping (nDTW), success weighted by path length (SPL), trajectory length (TL), and navigation error (NE). 
An episode is deemed successful if the agent stops within 3 meters of the goal in the VLN-CE environment, and within 2 meters in real-world environments. These thresholds ensure a fair comparison across different settings, balancing both navigational precision and room size.

\subsection{Comparison on Simulated Environment}
Table~\ref{table_performance_comparison} compares the performance of our proposed Open-Nav method against mainstream VLN methods. To ensure a fair comparison, all methods were evaluated using the same waypoint predictor, allowing us to isolate and focus on the performance of the navigator itself.
The first four methods listed in the table utilize supervised learning training methods, especially RecBERT~\cite{dcvln}, BEBbert~\cite{an2023bevbert} and ETPNav~\cite{An2023ETPNavET}, which benefit from pre-training and fine-tuning on in-domain datasets.
In the no-training category, we compared Open-Nav with several baselines: a navigator that randomly selects waypoint (Random), LXMERT~\cite{dcvln} trained without task-specific data, and other LLM-based methods such as DiscussNav~\cite{Long2023DiscussBM} that uses GPT-4 as a navigator. 
We can see that Open-Nav achieves superior results and outperforms DiscussNav in key metrics such as SR (5$\uparrow$) and SPL (2.39$\uparrow$).
In addition, we also evaluated our method using GPT-4 as a navigator, which indeed improves the performance such as SR and SPL, while our method using open-source LLM still achieves competitive performance.

\begin{table}[t]
\caption{Comparison on Simulated Environment R2R-CE dataset}
\label{table_performance_comparison}
\vspace{-15pt}
\begin{center}
\resizebox{\linewidth}{!}{
\begin{tabular}{l|cccccc}
\toprule
\textbf{Method} & \textbf{TL} & \textbf{NE}$\downarrow$ & \textbf{nDTW}$\uparrow$ & \textbf{OSR}$\uparrow$ & \textbf{SR}$\uparrow$ & \textbf{SPL}$\uparrow$ \\
\midrule
\multicolumn{7}{c}{\textbf{Supervised Learning}} \\
\midrule
CMA\cite{dcvln}& 11.08 & 6.92 & 50.77 & 45 & 37 & 32.17 \\
RecBERT\cite{dcvln}         & 11.06 & 5.8  & 54.81 & 57 & 48 & 43.22 \\
BEVBert\cite{an2023bevbert}               & 13.63 & 5.13 & 61.40 & 64 & 60 & 53.41 \\
ETPNav\cite{An2023ETPNavET}                 & 11.08 & 5.15 & 61.15 & 58 & 52 & 52.18 \\
\midrule
\multicolumn{7}{c}{\textbf{No Training}} \\
\midrule
Random                 & 8.15  & 8.63  & 34.08 & 12 & 2  & 1.50 \\
LXMERT\cite{dcvln} & 15.79 & 10.48 & 18.73 & 22 & 2  & 1.87 \\
DiscussNav-GPT4\cite{Long2023DiscussBM}             & 6.27 & 7.77  & 42.87 & 15 & 11 & 10.51 \\
\textbf{Open-Nav-Llama3.1(Ours)}         & 8.07  & \textcolor{blue}{\textbf{7.25}}  & \textcolor{blue}{\textbf{44.99}} & \textcolor{blue}{\textbf{23}} & \textcolor{blue}{\textbf{16}} & \textcolor{blue}{\textbf{12.90}} \\
\textbf{Open-Nav-GPT4(Ours)}         & 7.68  & \textbf{6.70}  & \textbf{45.79} & \textbf{23} & \textbf{19} & \textbf{16.10} \\
\bottomrule
\end{tabular}}
\vspace{-10pt}
\end{center}
\end{table}

\begin{table}[t]
\caption{Comparison on Real-world Environments}
\label{real}
\vspace{-15pt}
\begin{center}
\resizebox{\linewidth}{!}{
\begin{tabular}{l|cc|cc|cc|cc}
\toprule
\multirow{2}{*}{\textbf{Method}}  & \multicolumn{2}{c|}{{{Office}}} & \multicolumn{2}{c|}{{{Lab}}} & \multicolumn{2}{c|}{{{Game Room}}} & \multicolumn{2}{c}{{\textbf{All}}} \\
\cline{2-9}
                & {NE}$\downarrow$  & {SR}$\uparrow$  & {NE}$\downarrow$  & {SR}$\uparrow$  & {NE}$\downarrow$  & {SR}$\uparrow$ & \textbf{NE}$\downarrow$  & \textbf{SR}$\uparrow$ \\
\midrule
CMA\cite{dcvln}             &   2.50   &    30  &  2.77    &   20   &   2.65   &    20  & 2.64 & 23\\
RecBERT\cite{dcvln}         &   2.97   &    20  &   2.77   &   {35}   &    {2.47} &    {25} & 2.74 & 27\\
BEVBert\cite{an2023bevbert}         &   2.64   &   25   &    3.11  &     15 &   2.92   &    20 &2.89&20 \\
\textbf{Open-Nav-Llama3.1(Ours)} &  {2.44}  &  {35}   &  {2.36}    &   {35}  &     2.38 & {35}   & \textbf{2.39} & \textbf{35 }\\
\bottomrule
\end{tabular}}
\vspace{-15pt}
\end{center}
\end{table}

\begin{figure*}[t]
   \centering
     \includegraphics[width=0.95\linewidth]{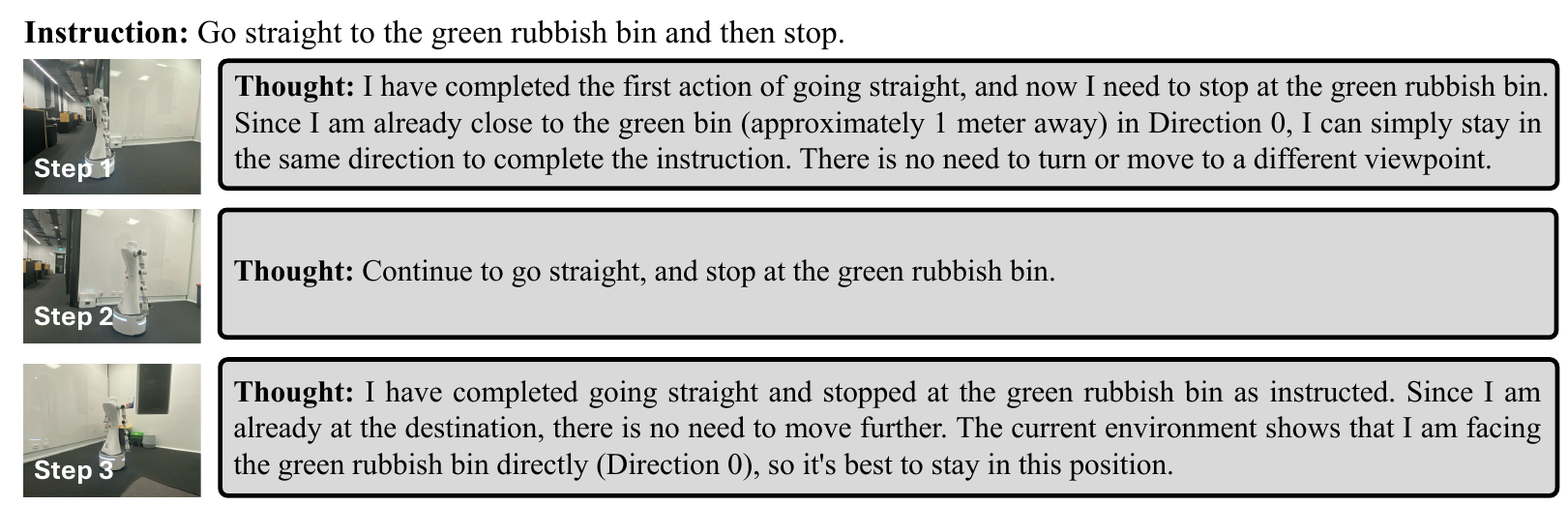}
    \caption{Visualization of our method Open-Nav in a real environment. The right side of the picture shows LLM's thoughts during navigation.}
    \label{fig:vis}
\end{figure*}

\subsection{Comparison on Real-world Environments}
To further evaluate the generalizability of methods in more challenging situations, we experiment with real-world environments.
As shown in Table~\ref{real}, we selected mainstream supervised training methods CMA~\cite{dcvln}, RecBERT~\cite{dcvln}, and BEVBert~\cite{an2023bevbert} to compare. Although these methods achieved good results in simulated environments, when these models were evaluated directly in real-world environments, these models were equivalent to no training setting. At this time, our Open-Nav achieved better performance, reaching SOTA in multiple scenarios. We speculate that this is because our Open-Nav has better generalization ability, since traditional supervised training models may not be able to recognize objects in real scenes because they are different from the objects seen in the training data.
We also provide some visual results of our method in Figure~\ref{fig:vis} in real-world environments, the right side shows the thoughts from LLM Navigator during the navigation process. It shows that Open-Nav well understands the task and the reasoning process is reasonable.

\subsection{Ablation Study}

\subsubsection{Different LLM on Instruction Comprehension}
\begin{figure}[t]
   \centering
   \vspace{-10pt}
     \includegraphics[width=0.9\linewidth]{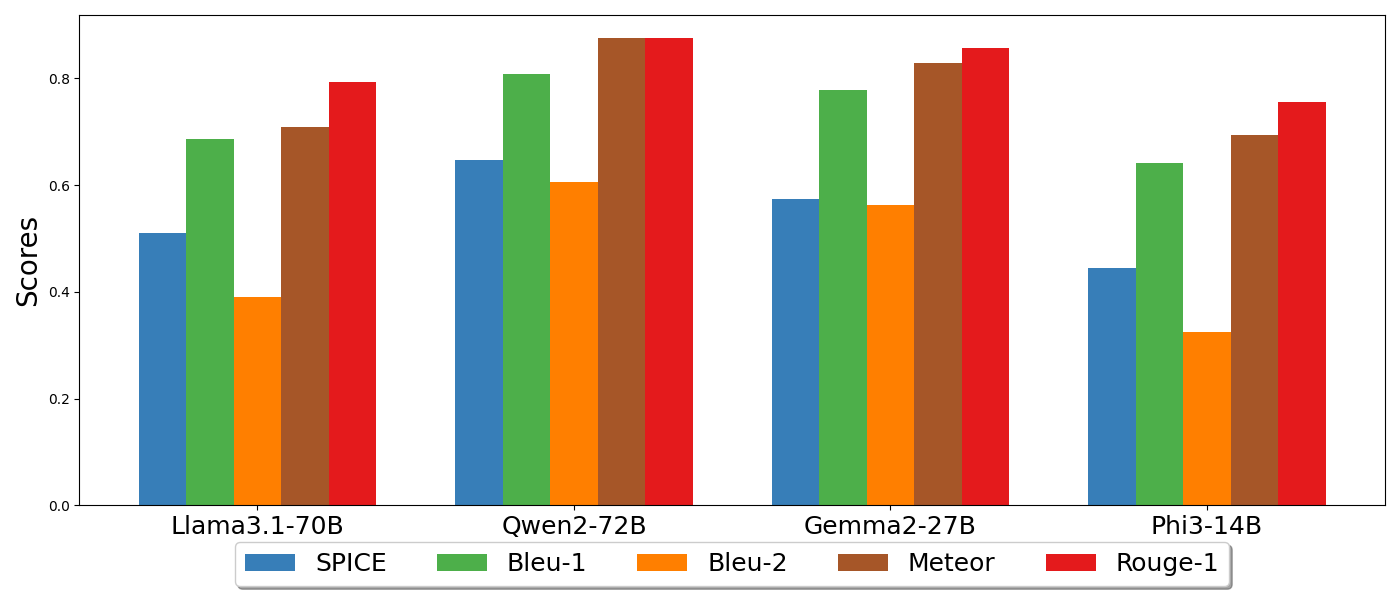}
     \vspace{-10pt}
    \caption{Performance of open-source LLMs on action decomposition.}
    \label{fig:bar_action}
    \vspace{-1pt}
\end{figure}

\begin{figure}[t]
   \centering
   \vspace{-10pt}
     \includegraphics[width=0.95\linewidth]{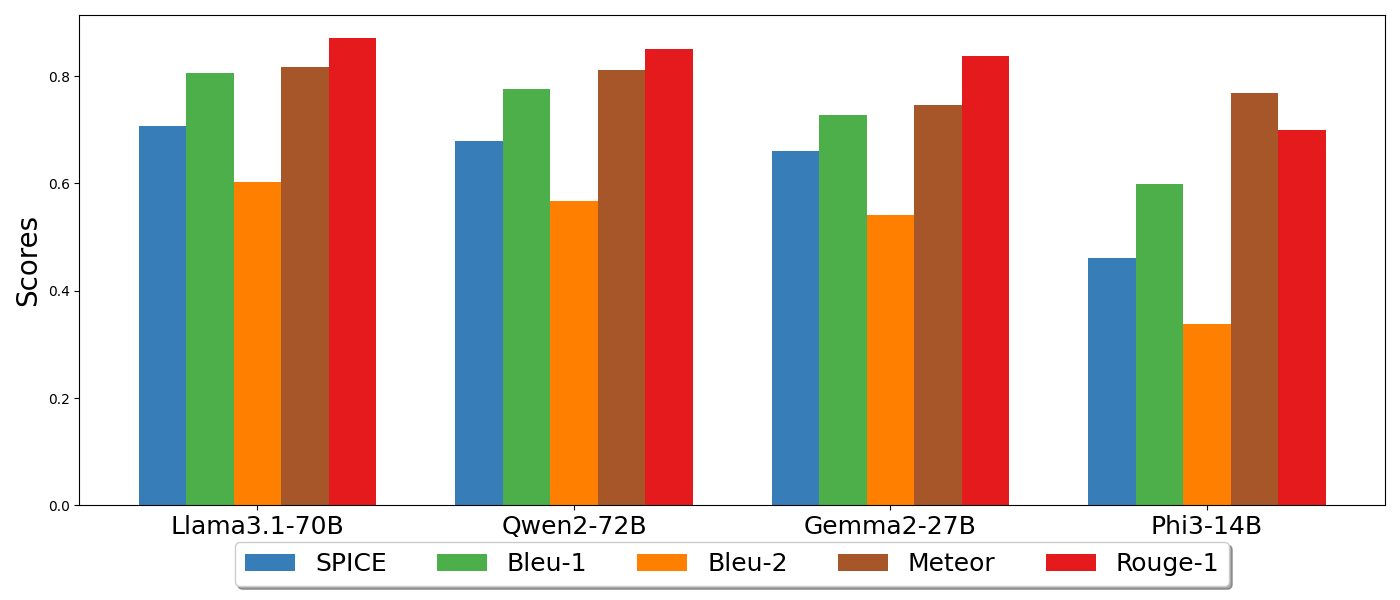}
     \vspace{-10pt}
    \caption{Performance of open-source LLMs on landmark extraction.}
    \label{fig:bar_landmark}
    \vspace{-10pt}
\end{figure}

We also evaluated the ability of different open-source LLMs in instruction understanding, with a particular focus on their ability to detect and decompose the actions and landmarks mentioned in the instructions. We first preprocessed the data using GPT-4 to label the required actions and landmarks in each instruction. Subsequently, these data were manually calibrated to obtain ground truth. To evaluate the performance of these LLMs in understanding instructions, we used four text-based metrics: SPICE~\cite{spice}, BLEU~\cite{Papineni2002BleuAM}, METEOR~\cite{Banerjee2005METEORAA}, and ROUGE~\cite{Lin2004ROUGEAP}.
The evaluation results are shown in Figure~\ref{fig:bar_action} and Figure~\ref{fig:bar_landmark}. It can be seen that among the four LLMs, Llama3.1-70B performs best in landmark extraction, while Qwen2-72B has the highest score in action decomposition. 

\begin{table}[t]
\caption{Performance of Different Open-Source LLMs on Navigation}
\label{table_llm}
\vspace{-15pt}
\begin{center}
\resizebox{\linewidth}{!}{
\begin{tabular}{l|cccccc}
\toprule
\textbf{Method} & \textbf{TL} & \textbf{NE}$\downarrow$ & \textbf{nDTW}$\uparrow$ & \textbf{OSR}$\uparrow$ & \textbf{SR}$\uparrow$ & \textbf{SPL}$\uparrow$ \\
\midrule
Llama3.1-70B             & 8.07  & 7.25  & \textbf{44.99} & \textbf{23} & \textbf{16} & \textbf{12.90} \\
Qwen2-72B   & 7.21 & 8.14 & 43.14 & \textbf{23} & 14  & 12.11 \\
Gemma-27B   & 8.41 & \textbf{6.76}  & 40.57 & 16 & 12 & 10.65 \\
Phi3-14B    & 8.47  & 8.53  & 33.64 & 8 & 5 & 3.81 \\
\bottomrule
\end{tabular}}
\vspace{-10pt}
\end{center}
\end{table}

\subsubsection{Different LLM on Navigation}
In addition to instruction understanding, we are more concerned about the decision-making ability of different LLMs when acting as navigators. Therefore, we also evaluated the navigation performance of these four LLMs. As shown in Table~\ref{table_llm}, although the performance of Llama3.1-70B and Qwen2-72B is comparable in the instruction understanding part, Llama3.1-70B is better in actual navigation, with a higher success rate. 

\section{Conclusion and future work}
In this paper, we present Open-Nav, a novel study that explores zero-shot VLN-CE tasks using open-source Large Language Models (LLMs). Unlike previous approaches that rely on expensive API-based models such as GPT-4, Open-Nav resorts to deploying open-source LLMs locally, addressing potential privacy issues and reducing operational costs.
Open-Nav adopts a spatial-temporal CoT approach from three viewpoints: instruction understanding, progress estimation, and decision making. 
To enhance scene perception, Open-Nav integrates RGB and depth data, leverages a spatial understanding model for spatial reasoning, and a fine-grained object recognition model for detailed object recognition. We conduct extensive experiments in both simulated and real-world environments, demonstrating that Open-Nav achieves competitive performance compared to GPT4-based approaches while maintaining low cost and privacy preservation. In the future, we plan to investigate methods to optimize the computational efficiency of open-source LLMs for navigation in real-world environments. 





\bibliographystyle{IEEEtran}
\bibliography{main}

\begin{thebibliography}{10}
\providecommand{\url}[1]{#1}
\csname url@rmstyle\endcsname
\providecommand{\newblock}{\relax}
\providecommand{\bibinfo}[2]{#2}
\providecommand\BIBentrySTDinterwordspacing{\spaceskip=0pt\relax}
\providecommand\BIBentryALTinterwordstretchfactor{4}
\providecommand\BIBentryALTinterwordspacing{\spaceskip=\fontdimen2\font plus
\BIBentryALTinterwordstretchfactor\fontdimen3\font minus \fontdimen4\font\relax}
\providecommand\BIBforeignlanguage[2]{{%
\expandafter\ifx\csname l@#1\endcsname\relax
\typeout{** WARNING: IEEEtran.bst: No hyphenation pattern has been}%
\typeout{** loaded for the language `#1'. Using the pattern for}%
\typeout{** the default language instead.}%
\else
\language=\csname l@#1\endcsname
\fi
#2}}

\bibitem{r2r}
P.~Anderson, Q.~Wu, D.~Teney, J.~Bruce, M.~Johnson, N.~S{\"{u}}nderhauf, I.~D. Reid, S.~Gould, and A.~van~den Hengel, ``Vision-and-language navigation: Interpreting visually-grounded navigation instructions in real environments,'' in \emph{IEEE Conf. Comput. Vis. Pattern Recog.}, 2018, pp. 3674--3683.

\bibitem{Krantz2021WaypointMF}
J.~Krantz, A.~Gokaslan, D.~Batra, S.~Lee, and O.~Maksymets, ``Waypoint models for instruction-guided navigation in continuous environments,'' \emph{Int. Conf. Comput. Vis.}, pp. 15\,142--15\,151, 2021.

\bibitem{vlnce}
J.~Krantz, E.~Wijmans, A.~Majumdar, D.~Batra, and S.~Lee, ``Beyond the nav-graph: Vision-and-language navigation in continuous environments,'' in \emph{Eur. Conf. Comput. Vis.}, A.~Vedaldi, H.~Bischof, T.~Brox, and J.~Frahm, Eds., 2020, pp. 104--120.

\bibitem{regretful}
C.~Ma, Z.~Wu, G.~AlRegib, C.~Xiong, and Z.~Kira, ``The regretful agent: Heuristic-aided navigation through progress estimation,'' in \emph{IEEE Conf. Comput. Vis. Pattern Recog.}, 2019, pp. 6732--6740.

\bibitem{Zhu0CL20}
F.~Zhu, Y.~Zhu, X.~Chang, and X.~Liang, ``Vision-language navigation with self-supervised auxiliary reasoning tasks,'' in \emph{IEEE Conf. Comput. Vis. Pattern Recog.}, 2020, pp. 10\,009--10\,019.

\bibitem{rcm}
X.~Wang, Q.~Huang, A.~{\c{C}}elikyilmaz, J.~Gao, D.~Shen, Y.~Wang, W.~Y. Wang, and L.~Zhang, ``Reinforced cross-modal matching and self-supervised imitation learning for vision-language navigation,'' in \emph{IEEE Conf. Comput. Vis. Pattern Recog.}, 2019, pp. 6629--6638.

\bibitem{Li2022EnveditEE}
J.~Li, H.~Tan, and M.~Bansal, ``Envedit: Environment editing for vision-and-language navigation,'' in \emph{IEEE Conf. Comput. Vis. Pattern Recog.}, 2022, pp. 15\,386--15\,396.

\bibitem{wang2023scaling}
Z.~Wang, J.~Li, Y.~Hong, Y.~Wang, Q.~Wu, M.~Bansal, S.~Gould, H.~Tan, and Y.~Qiao, ``Scaling data generation in vision-and-language navigation,'' in \emph{Int. Conf. Comput. Vis.}, 2023, pp. 12\,009--12\,020.

\bibitem{airbert}
P.~Guhur, M.~Tapaswi, S.~Chen, I.~Laptev, and C.~Schmid, ``Airbert: In-domain pretraining for vision-and-language navigation,'' in \emph{Int. Conf. Comput. Vis.}, 2021, pp. 1634--1643.

\bibitem{hamt}
S.~Chen, P.~Guhur, C.~Schmid, and I.~Laptev, ``History aware multimodal transformer for vision-and-language navigation,'' in \emph{Adv. Neural Inform. Process. Syst.}, 2021, pp. 5834--5847.

\bibitem{navillm}
D.~Zheng, S.~Huang, L.~Zhao, Y.~Zhong, and L.~Wang, ``Towards learning a generalist model for embodied navigation,'' in \emph{IEEE Conf. Comput. Vis. Pattern Recog.}, 2024, pp. 13\,624--13\,634.

\bibitem{Zhang2024NaVidVV}
J.~Zhang, K.~Wang, R.~Xu, G.~Zhou, Y.~Hong, X.~Fang, Q.~Wu, Z.~Zhang, and W.~He, ``Navid: Video-based vlm plans the next step for vision-and-language navigation,'' \emph{ArXiv}, vol. abs/2402.15852, 2024.

\bibitem{saycan}
M.~Ahn, A.~Brohan, N.~Brown, Y.~Chebotar, O.~Cortes, B.~David, C.~Finn, C.~Fu, K.~Gopalakrishnan, K.~Hausman, A.~Herzog, D.~Ho, J.~Hsu, J.~Ibarz, B.~Ichter, A.~Irpan, E.~Jang, R.~J. Ruano, K.~Jeffrey, S.~Jesmonth, N.~Joshi, R.~Julian, D.~Kalashnikov, Y.~Kuang, K.-H. Lee, S.~Levine, Y.~Lu, L.~Luu, C.~Parada, P.~Pastor, J.~Quiambao, K.~Rao, J.~Rettinghouse, D.~Reyes, P.~Sermanet, N.~Sievers, C.~Tan, A.~Toshev, V.~Vanhoucke, F.~Xia, T.~Xiao, P.~Xu, S.~Xu, M.~Yan, and A.~Zeng, ``Do as i can and not as i say: Grounding language in robotic affordances,'' in \emph{arXiv preprint arXiv:2204.01691}, 2022.

\bibitem{Zhou2023NavGPTER}
G.~Zhou, Y.~Hong, and Q.~Wu, ``Navgpt: Explicit reasoning in vision-and-language navigation with large language models,'' in \emph{AAAI}, 2023.

\bibitem{Long2023DiscussBM}
Y.~Long, X.~Li, W.~Cai, and H.~Dong, ``Discuss before moving: Visual language navigation via multi-expert discussions,'' \emph{IEEE International Conference on Robotics and Automation}, pp. 17\,380--17\,387, 2023.

\bibitem{Dai2023InstructBLIPTG}
W.~Dai, J.~Li, D.~Li, A.~M.~H. Tiong, J.~Zhao, W.~Wang, B.~A. Li, P.~Fung, and S.~C.~H. Hoi, ``Instructblip: Towards general-purpose vision-language models with instruction tuning,'' \emph{ArXiv}, vol. abs/2305.06500, 2023.

\bibitem{cot}
J.~Wei, X.~Wang, D.~Schuurmans, M.~Bosma, B.~Ichter, F.~Xia, E.~H. Chi, Q.~V. Le, and D.~Zhou, ``Chain-of-thought prompting elicits reasoning in large language models,'' in \emph{Advances in Neural Information Processing Systems}, 2022.

\bibitem{dcvln}
Y.~Hong, Z.~Wang, Q.~Wu, and S.~Gould, ``Bridging the gap between learning in discrete and continuous environments for vision-and-language navigation,'' in \emph{IEEE Conf. Comput. Vis. Pattern Recog.}, 2022, pp. 15\,418--15\,428.

\bibitem{Cai2024SpatialBotPS}
W.~Cai, Y.~Ponomarenko, J.~Yuan, X.~Li, W.~Yang, H.~Dong, and B.~Zhao, ``Spatialbot: Precise spatial understanding with vision language models,'' \emph{ArXiv}, vol. abs/2406.13642, 2024.

\bibitem{Zhang2023RecognizeAA}
Y.~Zhang, X.~Huang, J.~Ma, Z.~Li, Z.~Luo, Y.~Xie, Y.~Qin, T.~Luo, Y.~Li, S.~Liu, Y.~Guo, and L.~Zhang, ``Recognize anything: A strong image tagging model,'' \emph{ArXiv}, vol. abs/2306.03514, 2023.

\bibitem{Dorbala2022CLIPNavUC}
V.~S. Dorbala, G.~A. Sigurdsson, R.~Piramuthu, J.~Thomason, and G.~S. Sukhatme, ``Clip-nav: Using clip for zero-shot vision-and-language navigation,'' \emph{ArXiv}, vol. abs/2211.16649, 2022.

\bibitem{mattport}
A.~X. Chang, A.~Dai, T.~A. Funkhouser, M.~Halber, M.~Nie{\ss}ner, M.~Savva, S.~Song, A.~Zeng, and Y.~Zhang, ``Matterport3d: Learning from {RGB-D} data in indoor environments,'' in \emph{International Conference on 3D Vision}, 2017, pp. 667--676.

\bibitem{oaam}
Y.~Qi, Z.~Pan, S.~Zhang, A.~van~den Hengel, and Q.~Wu, ``Object-and-action aware model for visual language navigation,'' in \emph{Eur. Conf. Comput. Vis.}, 2020, pp. 303--317.

\bibitem{nvem}
D.~An, Y.~Qi, Y.~Huang, Q.~Wu, L.~Wang, and T.~Tan, ``Neighbor-view enhanced model for vision and language navigation,'' in \emph{{Proceedings of the ACM International Conference on Multimedia}}, 2021.

\bibitem{graph}
Y.~Hong, C.~R. Opazo, Y.~Qi, Q.~Wu, and S.~Gould, ``Language and visual entity relationship graph for agent navigation,'' in \emph{Adv. Neural Inform. Process. Syst.}, 2020.

\bibitem{speakerfollower}
D.~Fried, R.~Hu, V.~Cirik, A.~Rohrbach, J.~Andreas, L.~Morency, T.~Berg{-}Kirkpatrick, K.~Saenko, D.~Klein, and T.~Darrell, ``Speaker-follower models for vision-and-language navigation,'' in \emph{Adv. Neural Inform. Process. Syst.}, 2018, pp. 3318--3329.

\bibitem{envdrop}
H.~Tan, L.~Yu, and M.~Bansal, ``Learning to navigate unseen environments: Back translation with environmental dropout,'' in \emph{NAACL-HLT}, 2019, pp. 2610--2621.

\bibitem{li2023kerm}
X.~Li, Z.~Wang, J.~Yang, Y.~Wang, and S.~Jiang, ``Kerm: Knowledge enhanced reasoning for vision-and-language navigation,'' in \emph{IEEE Conf. Comput. Vis. Pattern Recog.}, 2023, pp. 2583--2592.

\bibitem{prevalent}
W.~Hao, C.~Li, X.~Li, L.~Carin, and J.~Gao, ``Towards learning a generic agent for vision-and-language navigation via pre-training,'' in \emph{IEEE Conf. Comput. Vis. Pattern Recog.}, 2020, pp. 13\,134--13\,143.

\bibitem{altr}
H.~Huang, V.~Jain, H.~Mehta, A.~Ku, G.~Magalh{\~{a}}es, J.~Baldridge, and E.~Ie, ``Transferable representation learning in vision-and-language navigation,'' in \emph{Int. Conf. Comput. Vis.}, 2019, pp. 7403--7412.

\bibitem{bertvln}
A.~Majumdar, A.~Shrivastava, S.~Lee, P.~Anderson, D.~Parikh, and D.~Batra, ``Improving vision-and-language navigation with image-text pairs from the web,'' in \emph{Eur. Conf. Comput. Vis.}, 2020, pp. 259--274.

\bibitem{qiao2023hop+}
Y.~Qiao, Y.~Qi, Y.~Hong, Z.~Yu, P.~Wang, and Q.~Wu, ``Hop+: History-enhanced and order-aware pre-training for vision-and-language navigation,'' \emph{IEEE Trans. Pattern Anal. Mach. Intell.}, 2023.

\bibitem{Chen_2022_DUET}
S.~Chen, P.-L. Guhur, M.~Tapaswi, C.~Schmid, and I.~Laptev, ``Think global, act local: Dual-scale graph transformer for vision-and-language navigation,'' in \emph{IEEE Conf. Comput. Vis. Pattern Recog.}, 2022.

\bibitem{An2023ETPNavET}
\BIBentryALTinterwordspacing
D.~An, H.~Wang, W.~Wang, Z.~Wang, Y.~Huang, K.~He, and L.~Wang, ``Etpnav: Evolving topological planning for vision-language navigation in continuous environments,'' \emph{IEEE Transactions on pattern analysis and machine intelligence}, vol.~PP, 2023. [Online]. Available: \url{https://api.semanticscholar.org/CorpusID:257985276}
\BIBentrySTDinterwordspacing

\bibitem{an2023bevbert}
D.~An, Y.~Qi, Y.~Li, Y.~Huang, L.~Wang, T.~Tan, and J.~Shao, ``Bevbert: Multimodal map pre-training for language-guided navigation,'' in \emph{Int. Conf. Comput. Vis.}, 2023, pp. 2737--2748.

\bibitem{wang2023gridmm}
Z.~Wang, X.~Li, J.~Yang, Y.~Liu, and S.~Jiang, ``Gridmm: Grid memory map for vision-and-language navigation,'' in \emph{Int. Conf. Comput. Vis.}, 2023, pp. 15\,625--15\,636.

\bibitem{song2022llm}
C.~H. Song, J.~Wu, C.~Washington, B.~M. Sadler, W.-L. Chao, and Y.~Su, ``Llm-planner: Few-shot grounded planning for embodied agents with large language models,'' \emph{arXiv preprint arXiv:2212.04088}, 2022.

\bibitem{qiao2023march}
Y.~Qiao, Y.~Qi, Z.~Yu, J.~Liu, and Q.~Wu, ``March in chat: Interactive prompting for remote embodied referring expression,'' in \emph{Int. Conf. Comput. Vis.}, 2023, pp. 15\,758--15\,767.

\bibitem{a2vnav}
P.~Chen, X.~Sun, H.~Zhi, R.~Zeng, T.~H. Li, G.~Liu, M.~Tan, and C.~Gan, ``A\({}^{\mbox{2}}\)nav: Action-aware zero-shot robot navigation by exploiting vision-and-language ability of foundation models,'' \emph{CoRR}, vol. abs/2308.07997, 2023.

\bibitem{zhan2024mc}
Z.~Zhan, L.~Yu, S.~Yu, and G.~Tan, ``Mc-gpt: Empowering vision-and-language navigation with memory map and reasoning chains,'' \emph{arXiv preprint arXiv:2405.10620}, 2024.

\bibitem{Long2024InstructNavZS}
Y.~Long, W.~Cai, H.~Wang, G.~Zhan, and H.~Dong, ``Instructnav: Zero-shot system for generic instruction navigation in unexplored environment,'' \emph{ArXiv}, 2024.

\bibitem{resnet}
K.~He, X.~Zhang, S.~Ren, and J.~Sun, ``Deep residual learning for image recognition,'' in \emph{IEEE Conf. Comput. Vis. Pattern Recog.}\hskip 1em plus 0.5em minus 0.4em\relax {IEEE} Computer Society, 2016, pp. 770--778.

\bibitem{Li2022BLIPBL}
J.~Li, D.~Li, C.~Xiong, and S.~C.~H. Hoi, ``Blip: Bootstrapping language-image pre-training for unified vision-language understanding and generation,'' in \emph{Int. Conf. Mach. Learn.}, 2022.

\bibitem{spice}
P.~Anderson, B.~Fernando, M.~Johnson, and S.~Gould, ``{SPICE:} semantic propositional image caption evaluation,'' in \emph{Eur. Conf. Comput. Vis.}, 2016, pp. 382--398.

\bibitem{Papineni2002BleuAM}
K.~Papineni, S.~Roukos, T.~Ward, and W.-J. Zhu, ``Bleu: a method for automatic evaluation of machine translation,'' in \emph{Annual Meeting of the Association for Computational Linguistics}, 2002.

\bibitem{Banerjee2005METEORAA}
S.~Banerjee and A.~Lavie, ``Meteor: An automatic metric for mt evaluation with improved correlation with human judgments,'' in \emph{IEEvaluation@ACL}, 2005.

\bibitem{Lin2004ROUGEAP}
C.-Y. Lin, ``Rouge: A package for automatic evaluation of summaries,'' in \emph{Annual Meeting of the Association for Computational Linguistics}, 2004.

\end{thebibliography}

\end{document}